\documentclass[sn-mathphys-num]{sn-jnl}

\usepackage{graphicx}%
\usepackage{multirow}%
\usepackage{amsmath,amssymb,amsfonts}%
\usepackage{amsthm}%
\usepackage{mathrsfs}%
\usepackage[title]{appendix}
\usepackage{xcolor}%
\usepackage{lmodern}
\usepackage{textcomp}%
\usepackage{manyfoot}%
\usepackage{booktabs}%
\usepackage{algorithm}%
\usepackage{algorithmicx}%
\usepackage{algpseudocode}%
\usepackage{listings}%
\usepackage{lineno}

\begin{document}

\title{Beyond Overconfidence: Model Advances and Domain Shifts Redefine Calibration in Neural Networks}


\author*[1,2,3]{\fnm{Achim} \sur{Hekler}}\email{achim.hekler@dkfz-heidelberg.de}

\author[1,2,3]{\fnm{Lukas} \sur{Kuhn}}

\author*[1,2,3,4]{\fnm{Florian} \sur{Buettner}}\email{florian.buettner@dkfz-heidelberg.de}

\affil[1]{\orgname{Goethe University Frankfurt}, \orgaddress{\city{Frankfurt}, \country{Germany}}}

\affil[2]{\orgname{German Cancer Consortium (DKTK)}, \orgaddress{\city{Frankfurt}, \country{Germany}}}

\affil[3]{\orgname{German Cancer Research Center (DKFZ)}, \orgaddress{\city{Heidelberg}, \country{Germany}}}

\affil[4]{\orgname{Frankfurt Cancer Institute}, \orgaddress{\city{Frankfurt}, \country{Germany}}}


\abstract{

Reliable uncertainty calibration of neural networks is crucial for safety-critical applications. 
Current calibration research has two major limitations: the exclusive evaluation of large web-scraped datasets and the lack of investigation of contemporary high-performance models with recent architectural and training innovations. 
To address this gap, we conducted a systematic investigation of different model generations on diverse datasets, revealing insights that challenge established calibration paradigms. Our results show that current-generation models consistently exhibit underconfidence in their in-distribution predictions - contrasting with the overconfidence typically reported in earlier model generations - while showing improved calibration under distribution shift.
Although post-hoc calibration techniques significantly improve in-distribution calibration performance, their effectiveness progressively diminishes with increasing distribution shift, ultimately becoming counterproductive in extreme cases. Critically, extending our analysis to four diverse biomedical imaging datasets using transfer learning highlights the limited transferability of insights from web-scraped benchmarks. In these domains, convolutional architectures  consistently achieve superior calibration compared to transformer-based counterparts, irrespective of model generation. 
Our findings underscore that model advancements have complex effects on calibration, challenging simple narratives of monotonic improvement, and emphasize the critical need for domain-specific architectural evaluation beyond standard benchmarks.

}




\maketitle

\section{} 
The deployment of deep neural networks (DNN) in real-world applications requires not only high predictive accuracy, but also precise and robust uncertainty quantification. This dual requirement is particularly critical in high-stakes domains such as medical diagnosis \cite{begoli_need_2019, kompa2021second}, autonomous driving \cite{tang2022}, and financial decision-making \cite{blasco2024}, where misaligned confidence estimates can lead to incorrect decisions with potentially severe or life-threatening  consequences. 

Model calibration offers a systematic framework for evaluating the reliability of a model's predictive confidence \cite{bayesian_binning, guo2017calibration}. In a well-calibrated model, confidence scores align closely with the true likelihood of correctness. For example, predictions made with 80 \% confidence should be correct approximately 80 \% of the time. This alignment between confidence and accuracy is essential for reliable decision-making in practical applications.

Significant progress has been made in the field of model calibration, shedding light on the challenges and limitations of neural networks. Early work, such as the seminal work by Guo et al.  \cite{guo2017calibration}, revealed that the neural networks of the time (e.g. ResNet or DenseNet) suffered from significant calibration issues, with confidence scores that poorly reflected their actual accuracy. 
In particular, the authors identified a systematic overconfidence, where models assign excessive confidence to incorrect predictions - a finding that has been corroborated by numerous subsequent studies \cite{thulasidasan_mixup_2019, hendrycks2021natural, lakshminarayanan_simple_2017, rahaman_uncertainty_2021}. 
Additionally, further studies    \cite{ovadia2019can, hendrycks_benchmarking_2019} emphasized that calibration performance deteriorates even further when models are exposed to distribution shifts, a scenario that is common in real-world applications. 

To address these challenges, researchers have developed post-hoc calibration techniques that refine a model's probability estimates \cite{guo2017calibration, zhang2020mixnmatchensemblecompositionalmethods, gupta_splines, tomani2022parameterized}.   These methods apply transformation functions to the model’s outputs, aligning confidence scores more closely with actual predictive accuracy. A key advantage of these techniques is their practicality: they improve calibration without requiring architectural changes or retraining, making them computationally efficient and well-suited for real-world deployment scenarios where resources are typically limited.

However, recent research has challenged these earlier conclusions. Minderer et al. \cite{minderer2021revisiting} demonstrated that state-of-the-art models of their time, such as Vision Transformers, exhibit significantly better inherent calibration properties and improved robustness to distribution shifts compared to their predecessors. These findings suggested a potential paradigm shift, implying that model advancements may inherently address some calibration challenges that were previously considered fundamental limitations of DNNs.

While these findings mark significant progress, two critical gaps remain in the study of model calibration. First, the rapid evolution of deep learning has introduced a new generation of high-performance models, incorporating innovations not only in architecture design \cite{Liu2022, bao2022beit, Fang2022EVAET}, but also pretraining strategies \cite{chen_empirical_2021, caron_emerging_2021, xie_self-supervised_2021}, and regularization techniques \cite{cubuk_randaugment_2020, yun_cutmix_2019, zhong_random_2020, touvron_going_2021}. While these advancements have set new benchmarks for predictive performance, their calibration properties remain largely unexplored. This raises a fundamental research question:
Can the relationship between model advancements and reduced calibration error, as established by Minderer et al., be confirmed for the latest generation of deep learning models that incorporate substantial architectural innovations alongside advanced training methodologies such as massive pretraining and sophisticated regularization techniques?

\begin{figure}
    \centering
    \includegraphics[width=\linewidth]{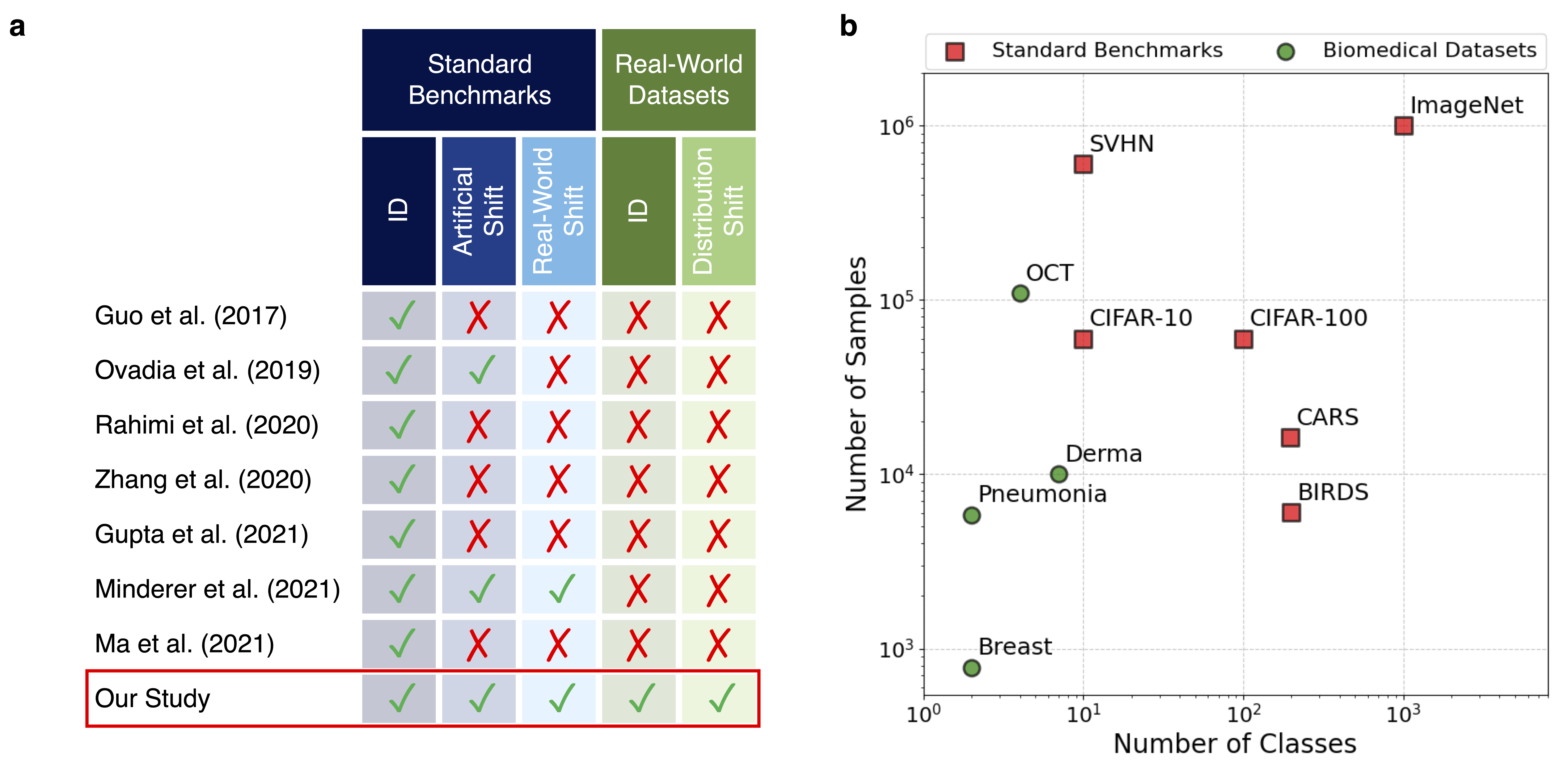}
    \caption{(a) Overview of a selection of important prior work on model calibration, illustrating the types of datasets and shifts analyzed. Most studies concentrate on the in-distribution setting of standard benchmarks, with some exploring artificial distribution shifts. However, only few address real-world shifted versions of these benchmarks. Our study uniquely integrates all these scenarios, including synthetic and real-world shifts, as well as real-world biomedical datasets, thereby filling a critical gap in calibration research.
    (b) Comparison of datasets, emphasizing  differences in size and classes. This study evaluates biomedical datasets (green circles) and standard web-scraped datasets (red squares), while previous calibration research exclusively relies on the latter ones. Web-scraped benchmarks  feature large sample sizes and numerous classes, whereas biomedical datasets typically have fewer samples and classes. 
    } 
    \label{fig:literature}
\end{figure}

Second, prior findings have primarily been based on web-scraped datasets such as ImageNet (Fig. \ref{fig:literature}a), where models can be optimized through  extensive pretraining and sophisticated regularization techniques.  The transferability of these findings to domains with practical constraints, such as biomedical applications, remains unclear. 
To address this, we conducted a systematic analysis using four biomedical datasets and applied best-practice transfer learning protocols. These datasets represent real-world scenarios that differ fundamentally from web-scraped collections: while ImageNet contains millions of images across one thousand well-balanced classes, biomedical datasets typically involve limited data, binary or few-class classification tasks, and imbalanced class distributions  (Fig. \ref{fig:literature}b). Given these differences, we investigate whether calibration properties established under optimal conditions hold in specialized domains with constrained data and training resources.

In summary, our analysis makes the following key contributions:

\begin{enumerate}

    \item We reveal that current-generation neural networks, such as  ConvNeXt, EVA, and BEiT, challenge established calibration principles in two key aspects: they exhibit systematic underconfidence in in-distribution predictions and show reduced calibration error under increasing distribution shift.

    \item We demonstrate a striking dichotomy in the effectiveness of post-hoc calibration methods on current-generation models: while these techniques significantly improve calibration for in-distribution predictions, their benefits diminish or even become detrimental under distribution shift.

    \item We show that calibration insights derived from common web-scraped datasets, such as ImageNet, exhibit limited transferability to specialized biomedical domains. 
    Our analysis uncovers a consistent architectural preference within biomedical imaging contexts: convolutional neural network architectures exhibit substantially superior calibration performance across all examined biomedical datasets compared to their transformer-based counterparts, while maintaining competitive predictive power. This architectural advantage persists irrespective of model generation.


    \item To facilitate rigorous, reproducible evaluation of post-hoc calibration methods within practical applications, we present a comprehensive open-access benchmark repository comprising meticulously curated model prediction datasets paired with standardized implementations of state-of-the-art calibration techniques. 
    
\end{enumerate}

\begin{figure} 
\centering 
\includegraphics[width=0.99\linewidth]{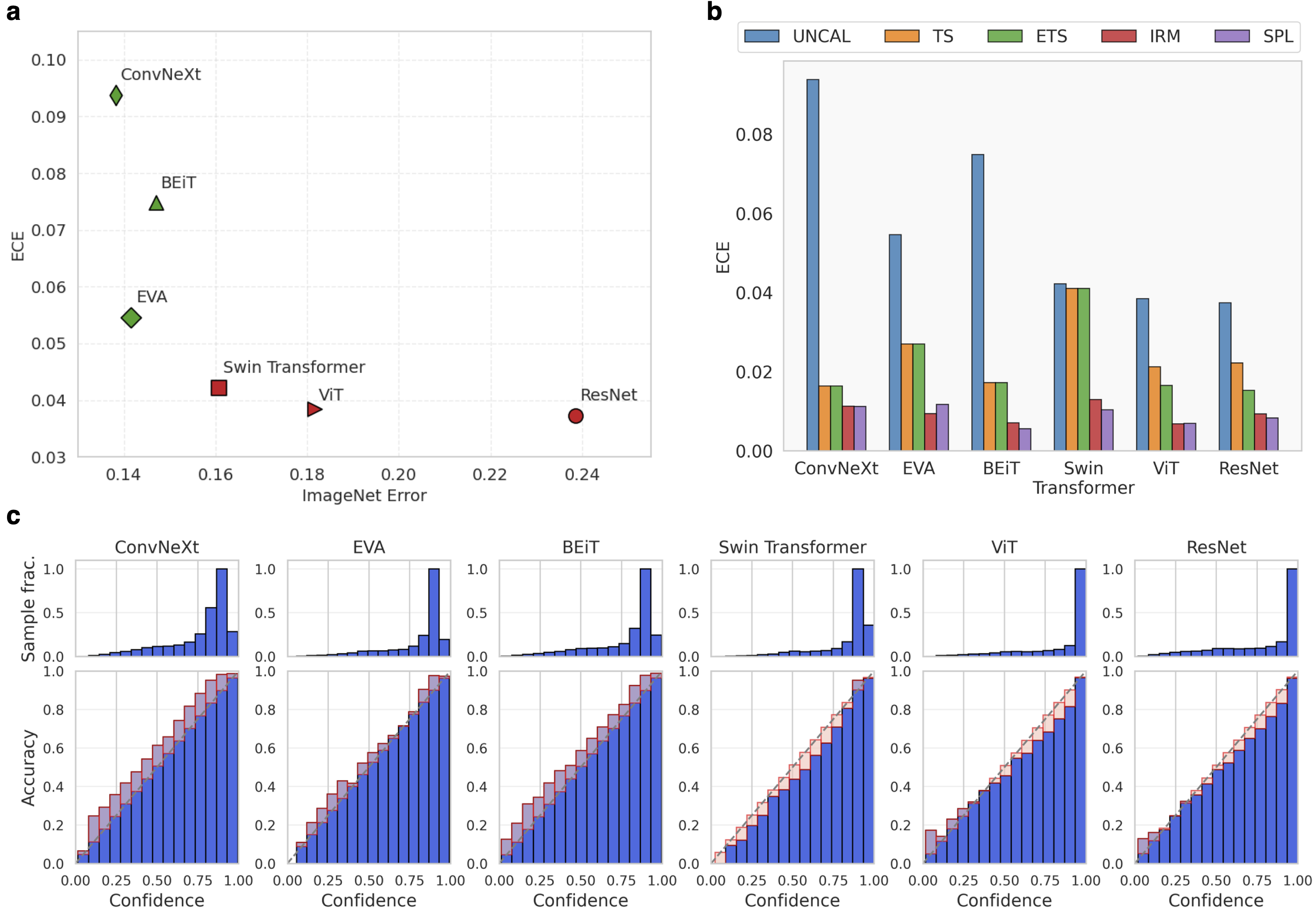} \caption{\textbf{Comprehensive analysis of calibration properties and the effectiveness of post-hoc calibration methods across different generations of neural networks on ImageNet.} (a) Scatter plot illustrating the inverse relationship between ImageNet classification error and Expected Calibration Error (ECE). Green markers represent current-generation models (ConvNeXt, EVA, BEiT), while red markers denote traditional models (Swin Transformer, ViT, ResNet). Despite their superior classification performance, current-generation models exhibit significantly higher calibration errors. (b) Comparative evaluation of post-hoc calibration methods (UNCAL: uncalibrated, TS: Temperature Scaling, ETS: Ensemble Temperature Scaling, IRM: accuracy-preserving Isotonic Regression, SPL: Spline Calibration), highlighting architecture-specific differences in effectiveness. (c) Reliability diagrams demonstrating the systematic underestimation of predictive confidence in current-generation models, contrasting with the overconfidence typically observed in traditional models.} 
\label{fig:ImageNet} 
\end{figure}

\section{Results} \label{sec:results}
We conducted a comprehensive empirical study to evaluate  both the inherent calibration properties  of neural networks and the effectiveness of post-hoc calibration techniques across a diverse range of architectures and application domains. Our analysis spans the evolution of deep learning models, from earlier convolutional and transformer-based networks, which had limited or no pretraining on large datasets, to modern hybrid designs that benefit from extensive pretraining and advanced regularization techniques. Specifically, we examine three current-generation models (ConvNeXt \cite{Liu2022}, BEiT \cite{bao2022beit}, and EVA \cite{Fang2022EVAET}) alongside three established models (Swin Transformer \cite{Liu_2021_ICCV}, Vision Transformer \cite{Dosovitskiy2020AnII}, and ResNet \cite{he2016deep}). This selection enables us to examine the evolution of calibration properties from earlier generations of models to the more advanced current-generation models that have emerged in recent years.

We begin our investigations with ImageNet \cite{ImageNet}, leveraging its scale and established role as a benchmark for evaluating fundamental calibration properties. To systematically assess robustness under distribution shift, we analyze calibration performance across a comprehensive set of 97 distribution-shifted datasets derived from ImageNet. This includes 95 synthetically generated datasets from ImageNet-C \cite{hendrycks_benchmarking_2019}, which offer a controlled environment for examining calibration properties across 19 types of common image corruptions at five severity levels. This setup enables a thorough evaluation of how calibration characteristics evolve under increasingly challenging conditions. Additionally, we extend our analysis to two real-world distribution shifts: ImageNet-V2 \cite{recht2019imagenet} and ImageNet-A \cite{hendrycks2021natural}.

We next assess, wether calibration insights derived from such web-scraped datasets transfer to the biomedical domain. To this end, we extend our analysis to four diverse biomedical datasets  \cite{al2020, tschandl2018, kermany2018} employing best-practice transfer learning techniques. These datasets encompass a variety of medical imaging modalities (X-ray, ultrasound, dermoscopy, and OCT), reflecting real-world clinical scenarios with diverse characteristics. The datasets range from smaller, specialized collections containing approximately 800 images to larger datasets with over 110,000 samples, covering both binary diagnostic tasks and multi-class classification challenges.

Throughout our main analysis, we use the Expected Calibration Error (ECE) with 15 equal-mass bins as our primary metric \cite{bayesian_binning}. Additionally, we report results for alternative ECE formulations, as well as complementary metrics such as the Brier score and negative log-likelihood in the Appendix. These additional evaluations consistently confirm the robustness of our findings across different calibration metrics. 

\subsection{Current-Generation In-Distribution Predictions Reveal an Inverse Relationship Between Accuracy and Calibration}

Our investigation on ImageNet reveals an unexpected inverse relationship between classification accuracy and calibration error in current-generation models (Fig. \ref{fig:ImageNet}a). While recent advancements in model design have significantly improved classification performance, we observe a notable decline in calibration quality compared to traditional models. This suggests a growing trade-off between these two critical aspects of model performance.
For instance, ConvNeXt achieves an impressive accuracy of 86.2 \% but exhibits poor calibration with an Expected Calibration Error (ECE) of 0.094. In contrast, the Vision Transformer, with a lower accuracy of 81.8 \%, demonstrates superior calibration, achieving an ECE of just 0.038.

Interestingly, these findings contradict the conclusions of Minderer et al. \cite{minderer2021revisiting}, who reported simultaneous improvements in accuracy and calibration as neural networks evolved. Our analysis reveals that recent innovations in model development have disrupted this trend, introducing a pronounced trade-off where gains in classification performance often come at the expense of calibration quality.

A closer examination using reliability diagrams (Fig. \ref{fig:ImageNet}c) highlights a distinct shift in the inherent calibration behavior of current-generation models. Unlike the systematic overconfidence observed in earlier studies \cite{guo2017calibration, hendrycks2021natural, lakshminarayanan_simple_2017, rahaman_uncertainty_2021}, current-generation models exhibit a tendency toward underconfidence in their in-distribution predictions. While this underconfidence increases overall calibration error, it may offer advantages in safety-critical domains, such as medical applications, where cautious predictions are often preferable to overconfident ones.

\subsection{Enhanced Responsiveness of Current-Generation Models to Post-hoc Calibration Techniques}
Given the increased ECE observed in current-generation models, we next evaluate the effectiveness of four established post-hoc calibration techniques:  Temperature Scaling (TS) \cite{guo2017calibration}, Ensemble Temperature Scaling (ETS) \cite{zhang2020mixnmatchensemblecompositionalmethods}, accuracy-preserving Isotonic Regression (IRM) \cite{zhang2020mixnmatchensemblecompositionalmethods}, and Spline calibration (SPL) \cite{gupta_splines}.

Our analysis demonstrates that current-generation models are significantly more responsive to post-hoc calibration techniques than traditional models (Fig. \ref{fig:ImageNet}b). For example, applying TS to ConvNeXt reduces its ECE from 0.094 to 0.016, representing a sixfold improvement in calibration quality. In comparison, TS reduces the ECE of the Vision Transformer from 0.038 to 0.021, achieving only a twofold improvement. Remarkably, the simplicity of TS is sufficient to align the calibration performance of current-generation models with that of traditional models, underscoring its effectiveness despite its minimal computational overhead.

Further analysis reveals other architecture-specific behaviors. For instance, the Swin Transformer shows limited responsiveness to both TS and ETS, suggesting that certain architectural constraints may inherently limit calibration improvements. Additionally, while ETS extends TS by incorporating ensemble-based adjustments, it provides minimal additional benefit for current-generation models compared to TS, despite its higher computational cost. This indicates that the added complexity of ETS may not always justify its use, particularly for architectures that already respond well to simpler methods.

On the other hand, methods like IRM and SPL consistently outperform both TS and ETS across all architectures, achieving superior calibration quality. These results establish IRM and SPL as reliable and effective solutions for in-distribution recalibration on ImageNet.

\begin{figure}
    \centering
    \includegraphics[width=0.99\linewidth]{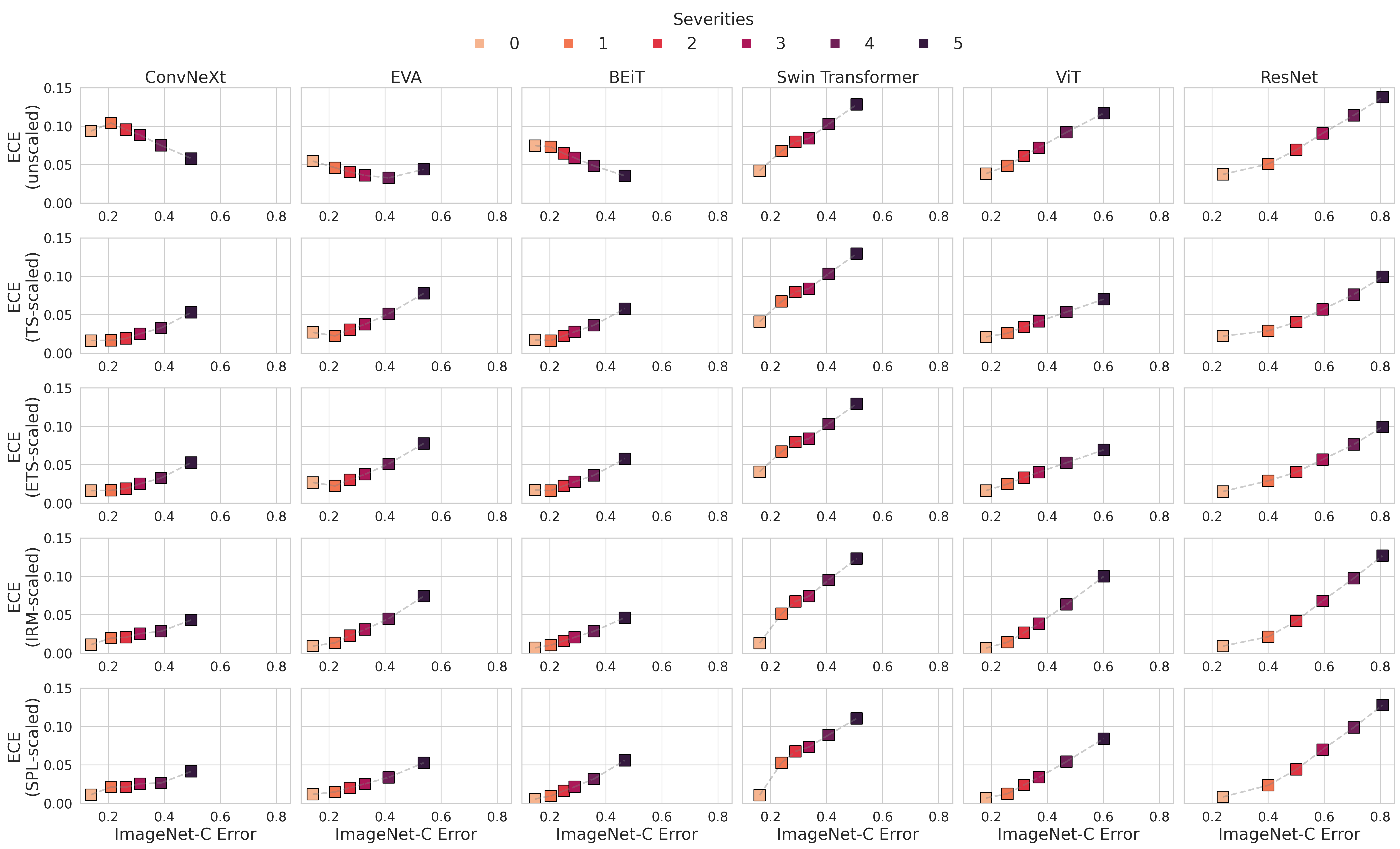}
    \caption{\textbf{Calibration performance under synthetic distribution shifts.} The Expected Calibration Error (ECE) and classification error are analyzed across severity levels on ImageNet-C, averaged over 19 perturbation types. The top row shows uncalibrated performance, where traditional architectures exhibit a monotonic increase in ECE with higher perturbation severity. In contrast, current-generation architectures display an inverse trend, with ECE decreasing as severity increases. Post-hoc calibration methods (e.g., Temperature Scaling, Ensemble Temperature Scaling, Isotonic Regression, and Splines) perform well under in-distribution conditions but degrade significantly under severe shifts, highlighting their limited robustness.}
    \label{fig:ImageNetC}
\end{figure}

\begin{figure}
    \centering
    \includegraphics[width=0.99\linewidth]{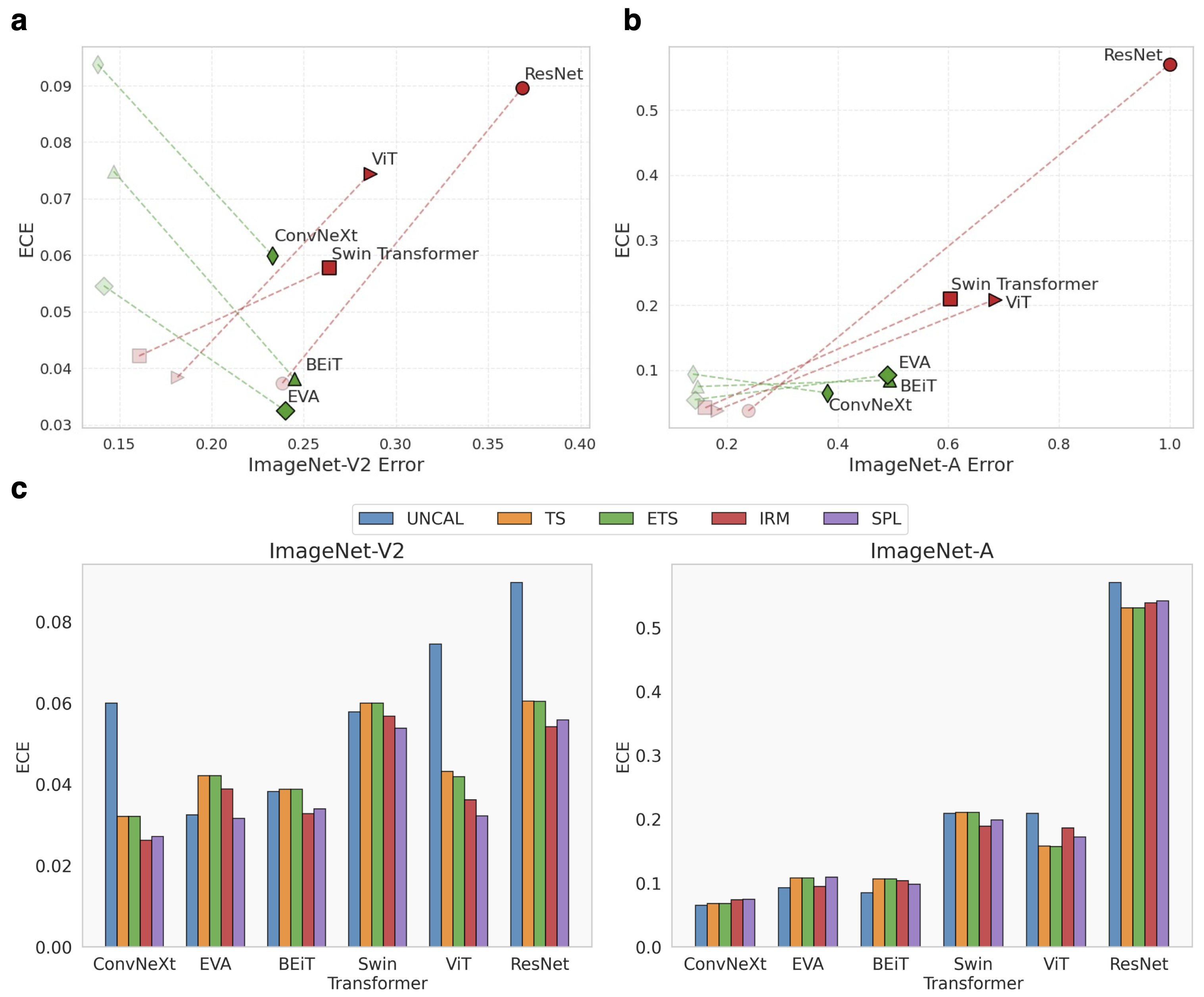}
    \caption{\textbf{Analysis of inherent calibration performance and effectiveness of post-hoc calibration methods under real-world distribution shift.} (a) Classification Error and Expected Calibration Error (ECE) on ImageNet-V2, comparing current-generation models (green markers) with traditional models (red markers). Grayed-out markers indicate corresponding in-distribution performance on ImageNet. Current-generation models demonstrate substantially lower ECE values compared to their in-distribution results, while traditional models exhibit increased ECE under distribution shift. (b) Classification Error and ECE across the considered models on ImageNet-A. In contrast to traditional models (green markers), which exhibit the expected substantial degradation in calibration performance under distribution shift, current-generation models (red markers) demonstrate remarkable calibration stability despite the challenging nature of the severe distribution shift. (c) Effectiveness of post-hoc calibration methods under distribution shift. The calibration performance of current-generation models deteriorates post-adjustment, with observed ECE values exceeding their uncalibrated baseline values.}
    \label{fig:ImageNetV2}
\end{figure}

\begin{figure}
    \centering
    \includegraphics[width=0.85\linewidth]{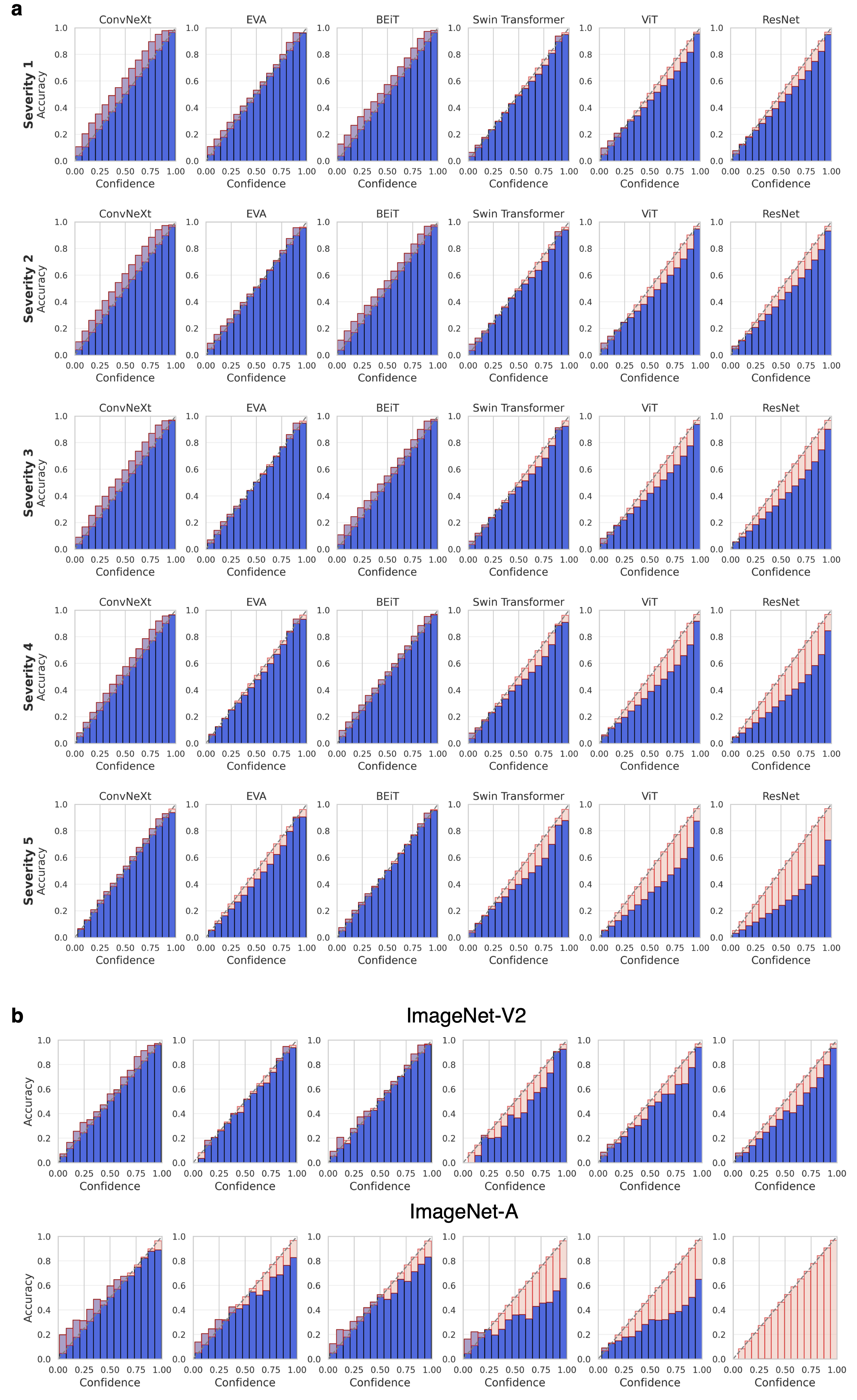}
    \caption{Reliability diagrams for (a) artificial distribution shift of ImageNet-C (averaged over the 19 corruption types)  and (b) real-world distribution shift of ImageNet-V2 and ImageNet-A. }
    \label{fig:reliabilityDiagramDistributionShift}
\end{figure}
\subsection{Current-Generation Models Demonstrate Improved Robustness Against Distribution Shift}

To systematically investigate calibration robustness under distribution shift, we first analyze model calibration on synthetic corruptions using ImageNet-C \cite{hendrycks_benchmarking_2019}. 


Our analysis reveals an interesting behavior of current-generation neural network calibration (Fig. \ref{fig:ImageNetC}). Previous studies consistently reported that both classification and calibration performance degrade as  distribution shift severity increases \cite{ovadia2019can, minderer2021revisiting, hendrycks_benchmarking_2019}. While this trend is confirmed  in our experiments with traditional models, current-generation models exhibit a more nuanced picture. Building on our observation that these models are inherently underconfident, we find that as corruption severity increases, their classification accuracy declines as expected, but their ECE values decrease. This behavior aligns with the established understanding that distribution shift tends to increase model confidence relative to accuracy, which in this case partially compensates for the initial underconfidence.

This architectural divide becomes even more apparent when examining real-world distribution shifts (Fig. \ref{fig:ImageNetV2}).
ImageNet-V2 and ImageNet-A represent distribution shifts of varying magnitudes, with ImageNet-A introducing substantially more challenging variations, as evidenced by the substantial drop in classification performance observed across all architectures.
On ImageNet-V2, current-generation models exhibit a reduction in ECE values, with improvements ranging from $\triangle$ECE = -0.022 to -0.037. In contrast, traditional models demonstrate an increase in ECE values, ranging from $\triangle$ECE = +0.016 to +0.051, aligning with previously reported trends in the literature. 
ImageNet-A's challenging distribution shift further amplifies the distinct calibration behaviors between traditional and current-generation architectures (Fig. \ref{fig:ImageNetV2}b).  Traditional models experience significant degradation in calibration performance, while current-generation models demonstrate remarkable stability in ECE values. Notably, ConvNeXt even achieves marginal improvements despite the highly challenging nature of the shift. 

Analysis of reliability diagrams across both synthetic and real-world shifts reveal consistent patterns (Fig. \ref{fig:reliabilityDiagramDistributionShift}). Distribution shifts generally increase model confidence relative to accuracy across all architectures, but with varying magnitudes.  For example, traditional models transition from overconfidence to extreme overconfidence as the severity of the distribution shift increases, a pattern observed in both synthetic and real-world scenarios. In contrast, current-generation models, which typically demonstrate underconfidence in in-distribution cases, shift towards a more calibrated state or exhibit slight overconfidence. Notably, ConvNeXt maintains its underconfident prediction patterns even under severe distribution shifts, whether in synthetic or real-world contexts.
Additionally, it is important to note that the magnitude of the confidence shift is model-dependent. Current-generation models exhibit only a minimal shift toward overconfidence, whereas the shift observed in traditional models is more pronounced.

\subsection{Post-hoc Calibration Fails Under Severe Distribution Shifts}
We next evaluate the performance of post-hoc calibration methods under varying levels of distribution shift. Our  findings reveal a critical dependency between the effectiveness of recalibration  and the severity of the shift, as illustrated in Fig.   \ref{fig:ImageNetC}.

 While post-hoc calibration methods significantly improve  calibration for current-generation models under in-distribution conditions and mild corruptions (severity levels 1-2), their effectiveness diminishes sharply as the severity of the distribution shift increases. 
 
 Notably, we observe that under sever distribution shifts, post-hoc calibration methods can degrade calibration performance to levels worse than those of uncalibrated models. While this negative interaction between post-hoc calibration and severe distribution shifts has received limited attention in the literature, it is particularly evident  in the EVA model.
 At severity levels 4 and 5, all calibration techniques - ranging from simple TS to more advanced methods - consistently yield ECE values that exceed those of the uncalibrated baseline. For instance, TS  results in a $\triangle$ECE of +0.018 at severity level 4 and $\triangle$ECE = +0.034 at severity level 5, highlighting the potential risks of applying recalibration techniques under extreme distribution shifts.
 
 To validate these findings, we extend our analysis to real-world distribution shifts using the ImageNet-V2 and ImageNet-A datasets (see Fig. \ref{fig:ImageNetV2}c). For the ConvNeXt model, post-hoc calibration methods demonstrate strong performance on ImageNet-V2, achieving significant reductions in ECE. However, as the distribution shift becomes more severe on ImageNet-A, the effectiveness of these methods declines sharply, with ECE values surpassing those of uncalibrated models. This trend mirrors the behavior observed under synthetic corruptions.

The EVA model exhibits a similar trend. Even under moderate shifts, such as those represented by ImageNet-V2, post-hoc calibration methods fail to improve calibration and instead result in higher ECE values compared to the uncalibrated baseline. This behavior is consistent with the trends observed under synthetic corruptions, where recalibration performance begins to degrade at lower severity levels.

In contrast, ResNet and ViT show more consistent improvements across both datasets, although the magnitude of improvement varies. The enhancements are more pronounced on ImageNet-V2 than on ImageNet-A, reflecting the increasing challenges posed by severe distribution shifts. Notably, the absolute ECE after post-hoc calibration remains lower for current-generation models compared to traditional ones, even under severe shifts.

Interestingly, the Swin Transformer exhibits negligible responsiveness to post-hoc calibration methods across both synthetic and real-world shifted datasets. This lack of responsiveness reaffirms the pattern observed in in-distribution cases, indicating that the Swin Transformer may possess specific design elements that hinder effective post-hoc recalibration. 

\begin{figure}
    \centering
    \includegraphics[width=0.88\linewidth]{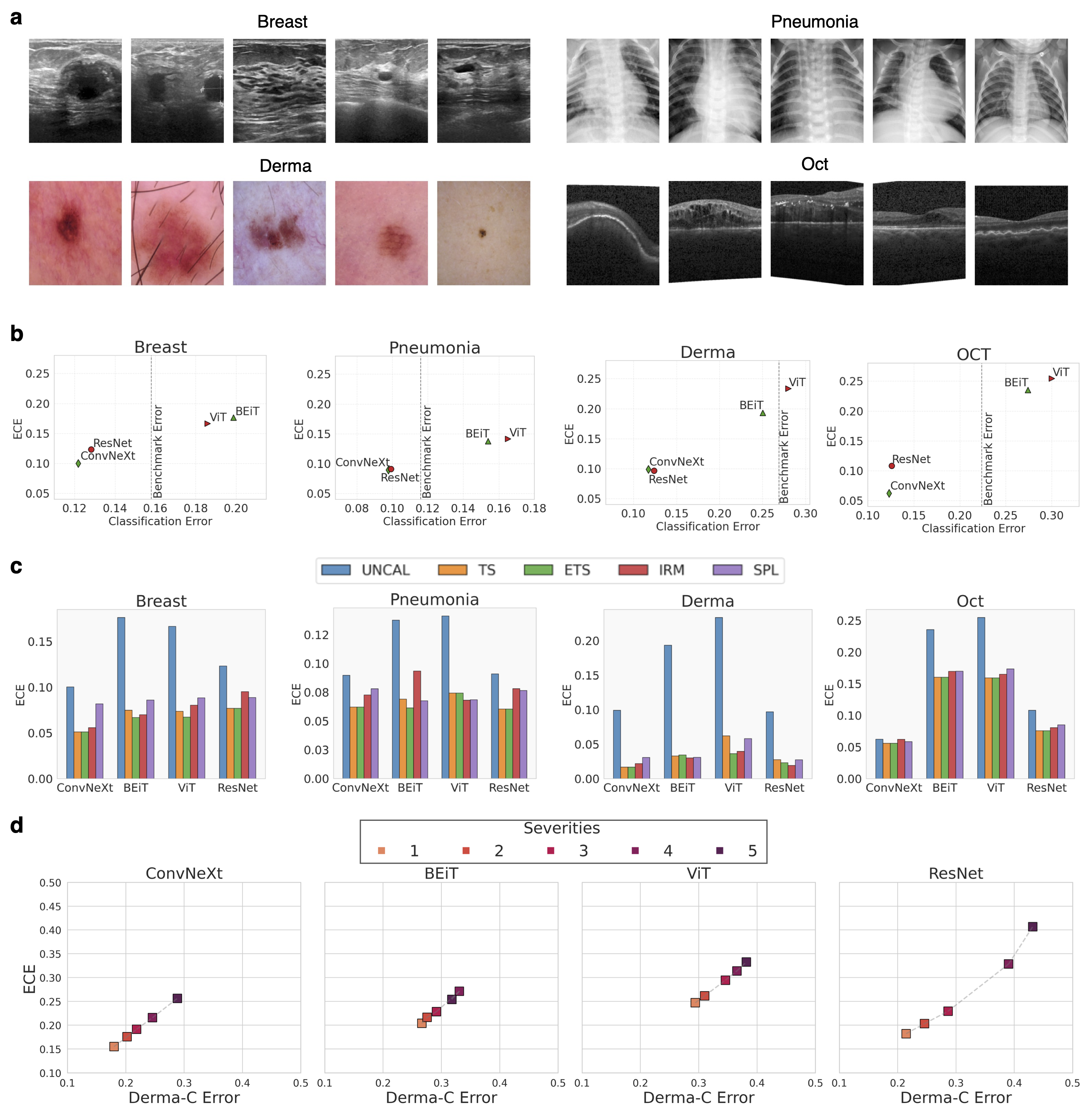}
    \caption{\textbf{Comparison of model calibration across different model architectures on four biomedical imaging datasets.} (a) Representative examples from four biomedical imaging datasets: breast ultrasound for cancer detection, chest X-rays for pneumonia diagnosis, dermoscopy images for skin lesion classification, and optical coherence tomography (OCT) for retinal disease diagnosis. Each dataset demonstrates the standardized nature of medical imaging, with consistent acquisition protocols and subtle diagnostic differences. (b) Scatter plots showing the relationship between classification error  and expected calibration error (ECE) for different model architectures across the four medical datasets. The dashed line represents the benchmark error reported in \cite{MEDMNISTv2} for the ResNet-50 architecture with image size 224. (c) Comparison of ECE values across different architectures (ConvNeXt, BEiT, ViT, ResNet) and different post-hoc calibration methods (indicated by colored bars) for each dataset. (d) Impact of distribution shift on classification error and ECE using 15 synthetic corruptions with five severities applied to the dermatological dataset. The ECE values are averaged over all corruption types.}
    \label{fig:MedMNIST}
\end{figure}

When comparing the performance of individual post-hoc calibration methods, an interesting trend emerges: the performance gap between temperature-based methods (TS and ETS) and more advanced techniques, such as Isotonic Regression (IRM) and Spline Calibration (SPL), narrows as the severity of the distribution shift increases. In some cases, TS and ETS even outperform IRM and SPL under extreme shifts. For example, on ViT at severity level 5, TS and ETS achieve better calibration performance compared to both IRM and SPL.

\subsection{Transfer Learning to Biomedical Tasks Reveals Limited Transferability of ImageNet Insights}


Finally, we systematically investigate wether our insights derived from ImageNet-based web-scraped datasets can be transferred to the biomedical domain. 
To this end, we examine calibration performance across four distinct biomedical datasets that reflect typical classification tasks in biomedical imaging. 
The fundamental differences between biomedical imaging data and conventional datasets like ImageNet present unique challenges and characteristics (Fig. \ref{fig:MedMNIST}a). Unlike ImageNet's natural images, which exhibit wide variations in object presentation and background complexity, biomedical images are obtained through standardized acquisition protocols, resulting in highly consistent image properties. The principal challenge in biomedical image analysis shifts from distinguishing visually distinct categories to identifying subtle pathological variations that typically challenge non-experts. These domain-specific attributes raise important questions about whether calibration properties observed in general computer vision tasks translate effectively to specialized medical applications.

In line with established workflows in applied biomedical image classification \cite{kumar2022finetuning}, we employ transfer learning strategies using ImageNet-pretrained models. Our study evaluates four distinct architectural approaches: classical convolutional networks (ResNet), vision transformers (ViT), and their contemporary variants (ConvNeXt and BEiT). 
The fine-tuning protocol yields models with excellent classification performance, with our ResNet implementation notably exceeding previously published benchmarks across all four biomedical datasets (Fig. \ref{fig:MedMNIST}b) \cite{MEDMNISTv2}.

Our detailed analysis of in-distribution calibration reveals distinctive patterns that contrast sharply with our ImageNet findings. Notably, architectures founded on convolutional principles (ResNet and ConvNeXt) demonstrate consistent superiority in both accuracy and calibration metrics across all biomedical tasks (Fig. \ref{fig:MedMNIST}b). We attribute this enhanced performance to three key characteristics of convolutional architectures that particularly benefit medical imaging applications:
(1) their intrinsic capacity for multi-scale feature extraction, facilitating the detection of both fine-grained and broad diagnostic patterns; (2) their transferable foundational capabilities in edge detection and texture analysis, derived from ImageNet pre-training; and (3) their inherent bias toward local pattern recognition, which aligns naturally with standardized medical imaging protocols. In contrast, Vision Transformer architectures, despite their success with natural images, may face limitations in medical applications, primarily due to their tokenized image representation approach that may not transfer effectively to biomedical imaging contexts.

Next, we investigate the effectiveness of post-hoc calibration methods, which varies significantly across the investigated biomedical datasets (Fig. \ref{fig:MedMNIST}c). In particular, on the dermatological dataset, post-hoc calibration techniques achieve calibration improvements comparable to those observed on natural image datasets. However, these same methods demonstrate reduced effectiveness in breast cancer detection, optical coherence tomography (OCT), and pneumonia diagnosis tasks.
Notably, TS and ETS emerge as the most reliable calibration methods across all biomedical datasets, contrasting with our ImageNet findings where more complex methods like IRM and SPL showed superior results. This unexpected performance gap suggests that sophisticated calibration methods, originally developed for natural images, may not generalize effectively to medical imaging contexts. This finding emphasizes the importance of domain-specific evaluation rather than assuming universal applicability of post-hoc calibration techniques.

Finally, we examine the impact of distribution shift on model calibration using a systematic approach with 15 
synthetic corruptions, including both general image corruptions and domain-specific perturbations, applied to the dermatological dataset (Fig. \ref{fig:MedMNIST}d) \cite{disalvo2024medmnistc}. 

Notably, the transfer learning process on the biomedical dataset appears to restore the widely documented positive correlation between ECE and classification error, a pattern that extends across all architectures including ConvNeXt and BEiT. This finding suggests that the remarkable robustness to distribution shift observed in our ImageNet experiments is not inherent to architectural designs but rather stems from training methodology, particularly the extensive pretraining on vast amounts of similar data. 

Examining the relative performance under distribution shift, BEiT demonstrates superior robustness in terms of both accuracy degradation and calibration error increase across corruption types. However, when considering absolute ECE values, ConvNeXT emerges as the superior model, maintaining both higher accuracy and lower ECE values under distribution shift compared to all other architectures. Particularly noteworthy is ConvNeXt's substantial improvement over ResNet, suggesting that modern architectural innovations successfully enhance robustness while preserving the advantages of convolutional principles, also for real-world biomedical applications.

\section{Discussion}
Our comprehensive investigation into neural network calibration reveals several findings that challenge current understanding of model calibration. These insights have significant implications for both research and practical applications.

Current-generation models exhibit an unexpected inverse relationship between classification and calibration error, but only on web-scraped datasets where extensive pretraining on similar large-scale datasets and comprehensive regularization and augmentation techniques are feasible. 
This contrasts with previous research \cite{minderer2021revisiting} and suggests that recent model advances influence how accuracy and calibration are balanced, often leading towards underconfident predictions rather than the overconfidence typically observed with traditional models.

Under distribution shift, current-generation models exhibit notable and distinctive behavior. While traditional models show increasing calibration error with increasing shift severity, current-generation models exhibit a decreasing ECE despite degrading classification accuracy. 
This pattern can be explained with prior research showing that distribution shift typically leads to increased model confidence relative to accuracy. In this specific case, this increased confidence partially offsets the model's initial underconfidence.

Regarding post-hoc calibration methods, our analysis highlights important limitations, particularly under severe distribution shifts. While these methods prove effective for in-distribution predictions, their performance can significantly deteriorate and even prove detrimental under substantial distribution shifts. This finding carries important implications for real-world deployments where distribution shifts are common as it suggests that architectural selection may be more critical than post-processing techniques in achieving robust calibration under distribution shift.

The transition to biomedical applications underscores the importance of domain-specific evaluation and reveals limits to the direct transferability of calibration insights derived solely from natural image domains. Our results demonstrate that convolutional architectures maintain superior performance in medical imaging contexts, suggesting that methods optimized for natural images may not directly translate to specialized domains. 

While our study provides evidence for these unexpected calibration phenomena in current-generation models, several important limitations should be considered. Our analysis primarily focuses on characterizing these phenomena rather than investigating their underlying causes. The observed systematic underconfidence could stem from multiple factors: architectural design choices, the scale and diversity of pretrainind datasets, or specific training protocols such as advanced data augmentation and regularization techniques. The interaction between these factors remains unexplored, as disentangling their individual contributions would require a comprehensive follow-up study involving large-scale training efforts, which are beyond the scope of this work. In particular, such a study would necessitate systematic ablations across architectural components, pretraining datasets, regularization techniques, and optimization strategies, requiring carefully controlled experiments to isolate the effects of each factor.

As recommendations for practioners, our comprehensive analysis suggests ConvNeXt as the preferred architecture for deployment scenarios requiring reliable model calibration. While ConvNeXt exhibits higher calibration errors for in-distribution predictions compared to traditional architectures like Vision Transformer, this calibration error can be effectively mitigated through simple post-hoc calibration techniques such as temperature scaling. 

Most significantly, ConvNeXt demonstrates superior robustness under distribution shift, maintaining reliable uncertainty estimates even as deployment conditions deviate from training data. This robustness, combined with state-of-the-art accuracy and the effectiveness of simple calibration techniques across various domains, makes ConvNeXt particularly suitable for real-world applications where both performance and reliable uncertainty quantification are critical.




\section{Methods}

\subsection{Problem Setup and Definitions}
In this paper, we focus on the calibration of neural networks trained on multi-class classification problems, where we observe a $D$-dimensional input variable $\textbf{x} \in \mathbf{R}^D$ and predict a categorical variable $\textbf{y} \in \{ 1,2, \dots, C\}$. The neural network outputs probabilistic predictions through a function $f$ that maps for every input instance $\textbf{x}$ a normalized probability vector $f(\textbf{x})$ assigning likelihoods to each of the $C$ possible classes. Mathematically, these predictions lie in the $(C-1)$ - dimensional probability simplex,  defined as 
\begin{equation*}
    \triangle = \left\{ p \in [0, 1]^C \, \bigg| \, \sum_{y=1}^C p_y = 1 \right\} .
\end{equation*} 

A model $f$ is considered well-calibrated when its outputs accurately reflect the level of predictive uncertainty. For instance, if we examine all inputs $\textbf{x}$ for which the model predicts $[ f(\textbf{x})]_y = 0.4$, we would anticipate that approximately 40 \% of these samples will actually be assigned the label $y$.

More formally, the model $f$ is calibrated if 
\begin{equation*}
    \forall p \in \triangle: \mathbb{P}(\textbf{y} = y \vert f(\textbf{x}) = p) = p_y \ .
\end{equation*} 

Throughout this paper, we will focus on a simplified, but more practical criterion, called top-label calibration. Here, the above equation does not have to be valid for all $p$,  but only for the most likely label, i.e.
\begin{equation}
    \forall p^* \in [ 0,1]: \mathbb{P}(\textbf{y} \in \arg \max f(\textbf{x}) \vert \max f(\textbf{x}) = p^*) = p^* \ .
    \label{eq: CalibrationCondition}
\end{equation} 

To quantify calibration performance, it is well-established to employ the Expected Calibration Error (ECE), which measures the expected discrepancy between the two sides of Eq. (\ref{eq: CalibrationCondition}) and is defined as

\begin{equation}
    \mathbb{E} [ \vert  p^* - \mathbb{P}(\textbf{y}  \in \arg \max f(\textbf{x}) \vert \max f(\textbf{x}) = p^*) \vert]
    \label{eq:ECE} \ .
\end{equation}

Due to the continuous-valued probability space, direct computation of Eq. (\ref{eq:ECE}) is intractable. Therefore, a discretization approach is typically employed by partitioning the prediction space into $m$ equally spaced  bins $B_1, \dots, B_m$. Given $n$ i.i.d. samples ${(x_i,y_i)}^{n}_{i=1}$ drawn from the joint distribution $\mathbb{P}(\textbf{x}, \textbf{y})$, we assign each $i \in \{ 1, \dots, n\}$ to a bin $B_j$ based on $\max f(x_i)$. 

Then, we compute for each bin $B_j$ the mean top-level confidence $\text{conf}(B_j) = \frac{1}{\vert B_j \vert} \sum_{i \in B_j} \max f(x_i)$ and the mean accuracy $\text{acc}(B_j) = \frac{1}{\vert B_j \vert} \sum_{i \in B_j} \mathbf{1}(\arg \max f(x_i) = y_i)$ and finally compute the Expected Caliration Error according to  

\begin{equation}
    \operatorname{ECE} = \sum_{j=1}^m \frac{|B_j|}{n} \left| \operatorname{acc}\left( B_j \right) - \operatorname{conf} \left( B_j \right) \right|.
\end{equation}
In addition to ECE, we quantify Brier score and the negative log likelihood as proper scoring rules, capturing both model calibration and model sharpness \cite{murphy1973new,popordanoska2024consistent}.  

\subsection{Datasets}
Our experimental evaluation encompasses both standard computer vision benchmarks and domain-specific biomedical imaging tasks, enabling a rigorous assessment of DNN calibration under varying data distributions.

For standard computer vision benchmarks, we utilize the \textbf{ImageNet} validation set as our primary benchmark, employing a random 90/10 split for testing and post-hoc calibration optimization, respectively.

To assess robustness under distribution shifts, we 
employ \textbf{ImageNet-C}, which extends the standard ImageNet dataset by incorporating 19 different types of corruptions, each at 5 severity levels. The corruptions include natural corruptions (e.g., fog, snow, frost), digital corruptions (e.g., JPEG compression, pixelation), and optical corruptions (e.g., motion blur, defocus blur). The graduated severity levels of these corruptions enable us to systematically evaluate model calibration quality as we move from in-distribution data to increasingly severe distribution shifts.
To optimize our post-hoc calibration methods, we utilize the dedicated calibration split from the original in-distribution ImageNet dataset. To maintain methodological integrity and prevent data leakage, we therefore carefully exclude these calibration images from all corresponding ImageNet-C corrupted versions during evaluation.

Although ImageNet-C offers systematic insights into calibration robustness, real-world applications frequently encounter distribution shifts that are more nuanced and complex than synthetic corruptions. To ensure practical relevance of our findings, we complement our synthetic corruption analysis with evaluation on two real-world distribution shifts:

\begin{enumerate}
    \item \textbf{ImageNet-V2} \cite{recht2019imagenet}, containing 10,000 temporally shifted samples collected using the original ImageNet sampling protocol.
    \item \textbf{ImageNet-A} \cite{hendrycks2021natural}, comprising  7,500 natural adversarial examples selected for their ability to induce misclassification in ResNet-50 architectures.
\end{enumerate}

For methodological consistency across all experiments, we optimize the post-hoc carlibration methods for both real-world distribution shifts using the in-distribution ImageNet calibration dataset.

To assess the generalization of post-hoc calibration methods to domain-specific tasks and real-world challenges, we incorporate four biomedical imaging datasets. These datasets encompass a wide spectrum of imaging modalities (microscopy, OCT, ultrasound), different dataset sizes (from 100 to $10^5$) and different classification tasks (binary and mult-class).  This heterogeneity in data characteristics facilitates a comprehensive and unbiased evaluation of model calibration across disparate biomedical imaging contexts. 

In detail, we investigate the following datasets:
\begin{itemize}
\item \textbf{Breast.} Our study incorporates the Breast Ultrasound Images (BUSI)  dataset \cite{al2020} consisting of 780 grayscale ultrasound images of breast tissue. This dataset presents a binary classification challenge between benign and malignant breast lesions, reflecting real-world clinical scenarios where limited data availability often constrains model development.
\item \textbf{Derma.}  
We utilize the HAM10000 dataset \cite{tschandl2018}, comprising 10,015 high-resolution dermatoscopic images categorized into seven distinct types of pigmented skin lesions. This dataset represents a comprehensive collection of dermatological conditions commonly encountered in clinical practice.
\item \textbf{OCT.} 
Our evaluation includes a large-scale retinal optical coherence tomography (OCT) dataset \cite{kermany2018} containing 109,309 high-resolution grayscale images. This dataset presents a four-class classification task for diagnosing retinal conditions.
\item \textbf{Pneumonia.} 
The pneumonia dataset \cite{kermany2018} consists of 5,586 grayscale pediatric chest radiographs. This collection presents a binary classification task for pneumonia detection, representing a common diagnostic challenge in clinical practice. 
\end{itemize}

For the four biomedical datasets, we adopt both the established train/validation/test splits and preprocessing pipeline from \cite{MEDMNISTv2}. This standardized approach ensures reproducibility and enables fair comparisons with existing and future studies in the field.

To systematically evaluate calibration robustness in biomedical applications, we conduct a distribution shift analysis on the dermatological dataset by applying both general image corruptions similar to ImageNet-C (e.g. Gaussian noise, blur, brightness variations) and domain-specific corruptions that simulate real-world clinical scenarios (e.g. black corners or characters on the image) \cite{disalvo2024medmnistc}. This comprehensive approach enables us to evaluate calibration performance under both general degradations and clinically relevant variations that practitioners might encounter. Example images illustrating these corruption types and their severity levels are provided in Appendix B.

\subsection{Model Selection and Training}
We evaluate calibration properties across a diverse set of neural network architectures spanning the evolution of computer vision models, from ResNet as a traditional CNN baseline to state-of-the-art designs like ConvNeXt, BEiT, and EVA. Our analysis encompasses pure convolutional architectures, transformer-based models, and hybrid approaches, enabling a comprehensive examination of how architectural paradigms influence calibration behavior. 

For our analysis on ImageNet, we use the following trained models from the timm library:
\begin{itemize}
    \item \textbf{ResNet-50} \cite{he2016deep}, which serves as our baseline architecture and is one of the most widely used models in biomedical AI research. For this study, we utilize the version trained on ImageNet-1k.
    \item \textbf{ViT-B/16} \cite{Dosovitskiy2020AnII}, which represents a pure transformer architecture that processes images as sequential patch embeddings. The Vision Transformer is known as a high-performing architecture and has been shown to exhibit good calibration properties in pevious study \cite{minderer2021revisiting}. In this work, we utilize the ViT-B/16 model, which is pretrained on ImageNet-21k and fine-tuned on ImageNet-1k.
    \item \textbf{Swin-S3-B} \cite{Liu_2021_ICCV}, which advances the transformer architecture through hierarchical feature representation and shifted window partitioning. This model is directly trained on ImageNet-1k.
    \item \textbf{BEiT-B/16} \cite{bao2022beit}, is a transformer-based image classification model designed to leverage self-supervised learning techniques. This model is pretrained on ImageNet-22k using masked image modelling. Following pretraining, the model is fine-tuned on ImageNet-22k and subsequently on ImageNet-1k.
    \item \textbf{EVA-S/14}  \cite{Fang2022EVAET}, which establishes new performance benchmarks through scaled transformer architectures. The model is pretrained on ImageNet-22k with masked image modelling and then fine-tuned on ImageNet-1k. 
    \item \textbf{ConvNeXt-B} \cite{Liu2022}, which bridges traditional and modern approaches by incorporating transformer-inspired design principles while preserving the inherent advantages of convolutional operations. In this study, we utilize a model, pretrained on ImageNet-22k and fine-tuned on ImageNet-1k.
\end{itemize}

For our biomedical imaging tasks, we train all model architectures through transfer learning, leveraging their ImageNet-pretrained weights. This approach is essential given the relatively small dataset size and has proven highly effective in many practical settings. Following established best practices for transfer learning, we implement a two-stage fine-tuning protocol:

For each architecture, we first adapt the model by replacing the final classification layer with a new one matching the target number of classes for each respective biomedical task. The training process then proceeds in two distinct phases:

\begin{enumerate}
\item \textit{Frozen Feature Extraction}: Initially, we freeze all layers except the newly added classification layer and train for several epochs, allowing the model to adapt its classification head to the new domain while preserving the pretrained feature extractors.
\item \textit{Progressive Fine-tuning}: Subsequently, we unfreeze the entire network and continue training with discriminative learning rates, applying smaller learning rates to early layers and gradually increasing them towards the final layers. This approach helps preserve useful low-level features while allowing task-specific adaptation of higher-level representations.
\end{enumerate}

The optimal hyperparameter for this process - including the number of epochs for each phase, learning rates for both frozen and unfrozen training, and the pooling type (Concat or Average Pooling) - were determined though cross-validation on the training set by optimizing the overall accuracy. 
In particular, the canonical validation sets of the biomedical datasets are exclusively used for optimization of the post-hoc calibration methods.

\subsection{Post-hoc Calibration Methods}
\label{subsec:posthoccalibrationmethods}
To systematically evaluate the effectiveness of post-hoc calibration techniques across different model architectures, we investigate several state-of-the-art methods that represent diverse approaches to post-hoc model calibration.

We examine the following calibration methods:
\begin{itemize}
    \item \textbf{Temperature Scaling (TS).} This post-hoc calibration method \cite{guo2017calibration} recalibrates network predictions using a single learned parameter T that rescales the model's pre-softmax logits. The parameter is optimized on validation data by minimizing a chosen loss function.
    TS's effectiveness stems from its ability to either soften overconfident predictions ($T > 1$) or sharpen underconfident ones ($T < 1$), while its single-parameter nature helps prevent overfitting.
\item \textbf{Ensemble Temperature Scaling (ETS).} Building upon TS, this method \cite{zhang2020mixnmatchensemblecompositionalmethods} constructs a weighted ensemble combining three probability distributions: temperature-scaled predictions, raw model outputs, and a uniform distribution over all classes. The weights and temperature parameter are jointly optimized on validation data, subject to the constraint that weights sum to one, offering enhanced calibration flexibility compared to standard TS.
\item \textbf{Isotonic Regression.} This data-efficient implementation \cite{zhang2020mixnmatchensemblecompositionalmethods} learns a single strictly monotonic calibration function by pooling prediction-label pairs across all classes. The approach minimizes squared loss between predictions and labels while maintaining model accuracy, avoiding the potential instability of class-wise calibration.
\item \textbf{Spline Calibration.} This non-parametric approach \cite{gupta_splines} learns continuous, piece-wise polynomial functions to recalibrate model outputs. The method partitions the probability space using fixed knot points and fits cubic splines between these points, creating a smooth calibration mapping that can capture complex relationships between predicted and true probabilities.
\end{itemize}

\subsection{Code Availability}
We implement and analyze the post-hoc calibration methods introduced in the previous section within our ModelTransformer Python package, which provides a unified framework inspired by scikit-learn's design principles. This package offers consistent interfaces for fitting and transforming data, enabling parameter estimation on validation datasets and subsequent application to test sets. 
The complete implementation is available at \url{https://github.com/MLO-lab/ModelTransformer/}.

The code used to generate all analysis and figures in this paper is publicly available at \url{https://github.com/MLO-lab/CalibrationStudy/}. This repository contains well-documented notebooks and scripts that enable complete reproduction of our experimental results and graphical representations.

\subsection{Data Availability}
The model outputs and associated data used in this paper are also publicly available at \url{https://doi.org/10.5281/zenodo.15229730}. 
This extensive collection of datasets enables complete reproduction of our calibration analysis and provides opportunities for researchers to conduct additional investigations beyond the scope of this work.

\appendix

\newpage
\section{Supplementary Material}

\section{Appendix A: Further Experimental Results}

\subsection{A1. Additional Calibration Metrics and Their Definitions}
To validate the robustness of our findings presented in the main paper, we extend our analysis of model calibration using different bin configurations and complementary calibration metrics. While the main analysis focused on Expected Calibration Error (ECE) with 15 bins, here we demonstrate that our conclusions  hold across the following calibration metrics:

\begin{enumerate}
    \item \textbf{Maximum Calibration Error (MCE)} measures the maximum difference between confidence and accuracy across all bins, providing insight into worst-case miscalibration:
    $$\text{MCE}= \max_j (p_j - c_j) ,$$ 
    where $p_j$ is the top-1 prediction accuracy in bin $j$ and $c_j$ is the average confidence of prediction in bin $j$.
    \item \textbf{Root Mean Square Error (RMSCE)} penalizes larger calibration errors more heavily than ECE by using squared differences:
    $$ \text{RMSCE} = \sqrt{\sum_j^m b_j (p_j - c_j)^2} ,$$
    where $b_j$ is the fraction of data points in bin $j$.
    \item \textbf{Root Brier Score (RBS)} measures the accuracy of probabilistic predictions:
    $$\text{BS} = \sqrt{\frac{1}{N}\sum_{i=1}^N\sum_{c=1}^C(p_{ic} - y_{ic})^2} , $$
    where $p_{ic}$ represents the predicted probability for class $c$ of sample $i$, and $y_{ic}$ is the corresponding one-hot encoded ground truth label.
    \item \textbf{Negative Log-Likelihood (NLL)} evaluates the quality of probabilistic predictions by measuring the likelihood of the true lables under the model's predicted distributions:
    $$\text{NLL} = - \frac{1}{N} \sum_{i=1}^N \log(p_{i, y_i}) , $$
    where $p_{i, y_i}$ is the predicted probability for the true class $y_i$ of sample $i$.
\end{enumerate}

\subsection{A2. Results for Inherent Model Calibration for Different ECE Configurations}

\begin{figure}[H]
    \centering
    \includegraphics[width=0.99\linewidth]{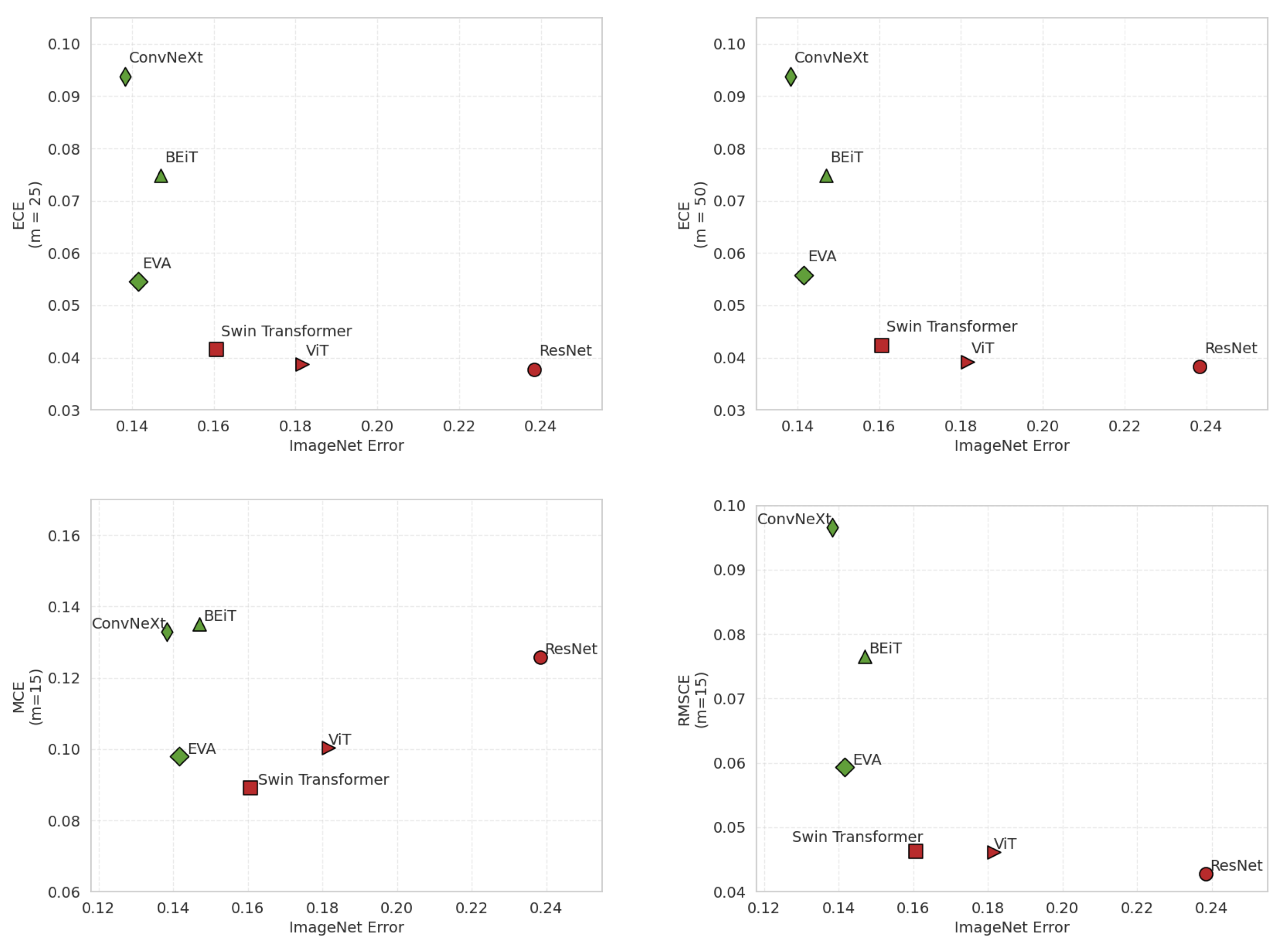}
    \caption{Scatter plot illustrating the inverse relationship between ImageNet cclassification error and ECE. Green markers represent current-generation models (ConvNeXt, EVA, BEiT), while red markers denote traditional models (Swin Transformer, ViT, ResNet). Scatter plot illustrating the inverse relationship between ImageNet classification error and ECE. Green markers represent current-generation models (ConvNeXt, EVA, BEiT), while red markers denote traditional models (Swin Transformer, ViT, ResNet). Furthermore, the results presented in the main body of the paper hold true for ECE across varying number of bins (m=25 and m=50) and for different ECE metrics, such as Maximum Calibration Error (MCE) and Root Mean Square Calibration Error (RMSCE).}
    \label{fig:appendix}
\end{figure}

\newpage

\subsection{A3. Effect of Bin Resolution on Reliability Diagram}

\begin{figure}[H]
    \centering
    \includegraphics[width=0.99\linewidth]{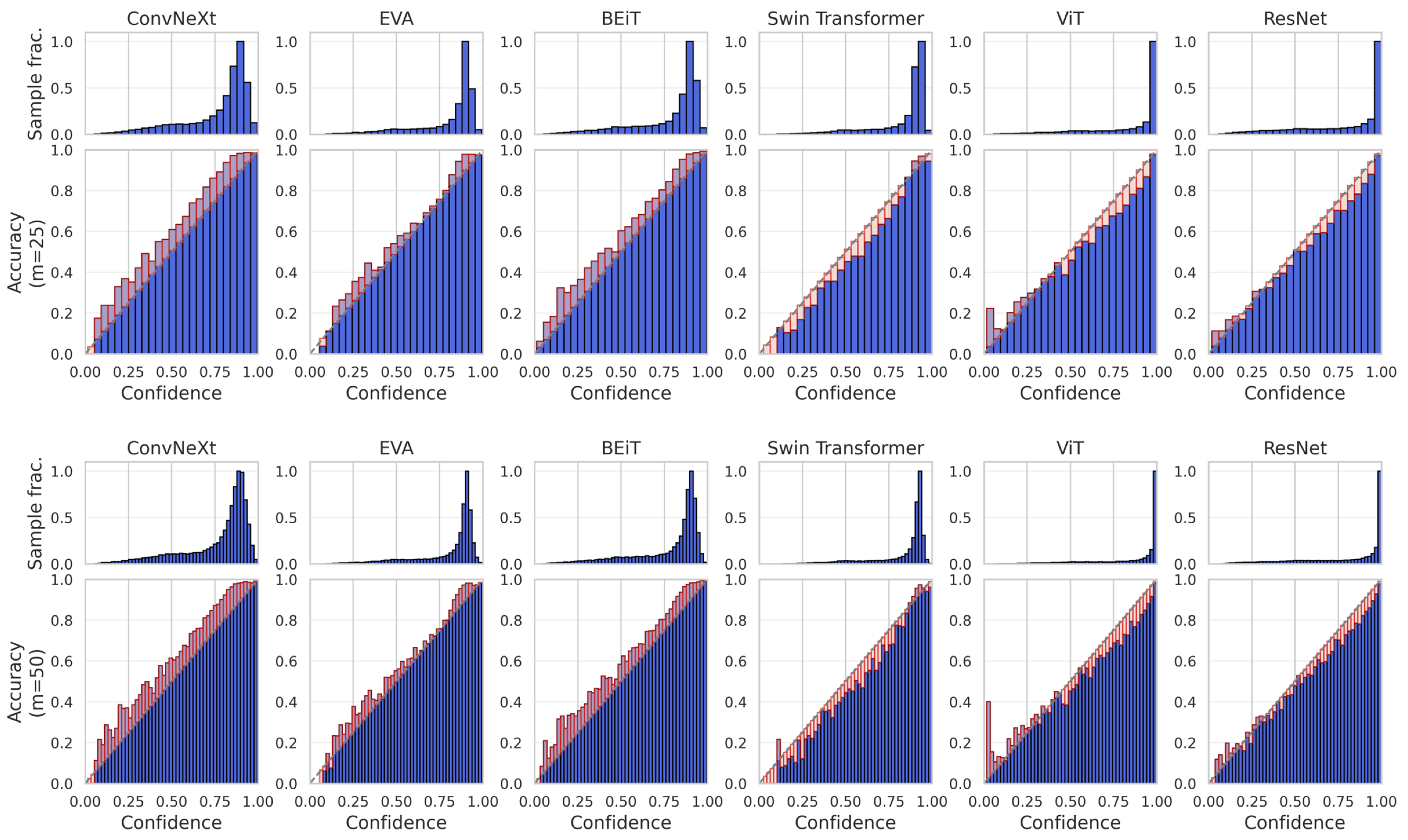}
    \caption{Reliability diagrams for current-generation and traditional models computed with different bin resolutions (25 and 50 bins). The diagrams demonstrate that the observed calibration patterns remain consistent across different bin counts, supporting the robustness of our findings.}
    \label{fig:reliability_imagenet_25_50}
\end{figure}

\newpage

\subsection{A4. Results for Post-hoc Calibration Techniques For Different Calibration Metrics}
\begin{figure}[H]
    \centering
    \includegraphics[width=0.99\linewidth]{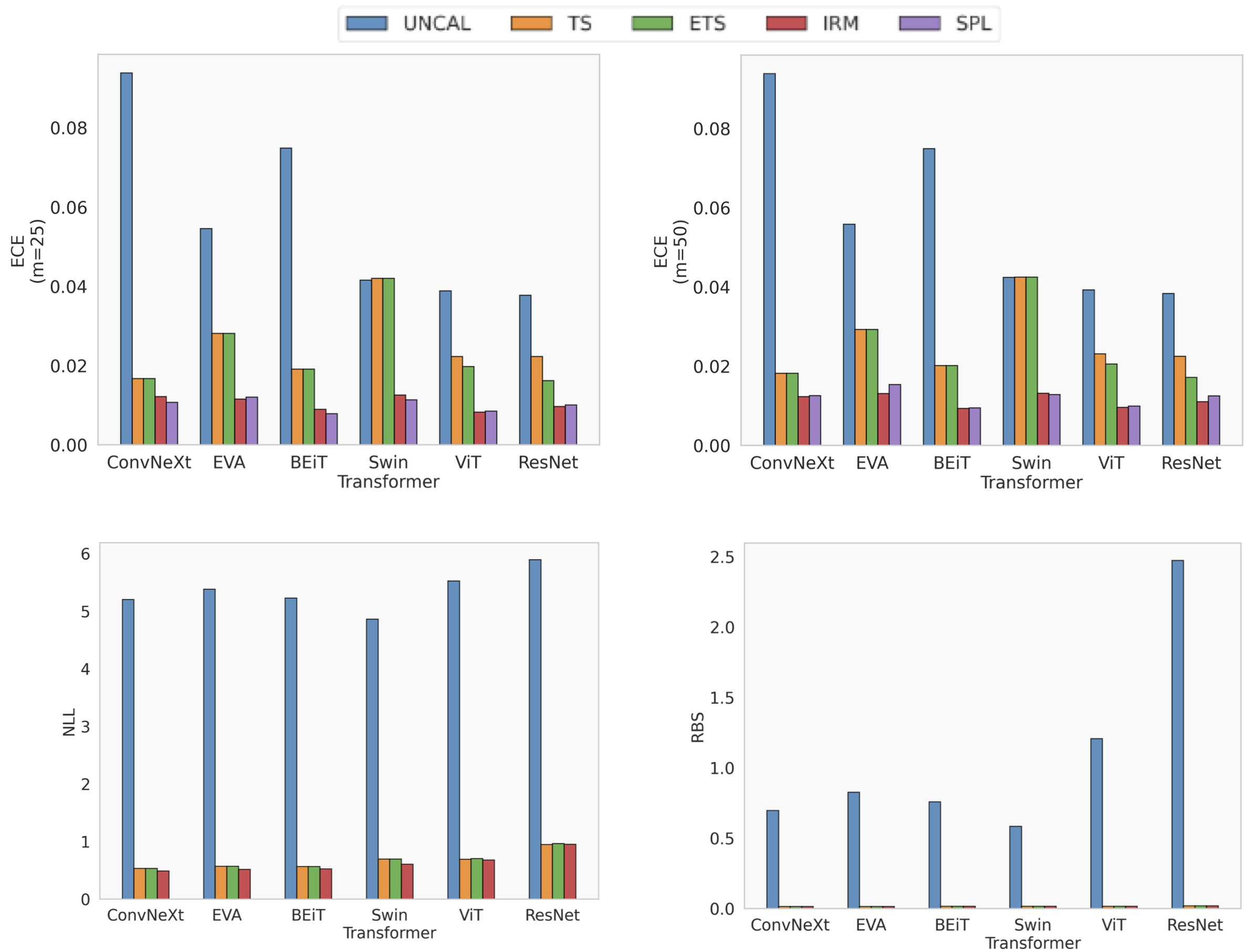}
    \caption{Comparison of post-hoc calibration effectiveness across multiple evaluation metrics: ECE with varying bin resolutions (25, and 50 bins), Root Brier Score (RBS), and Negative Log-Likelihood (NLL). }
    \label{fig:appendix_imageNet_posthoc}
\end{figure}

\newpage
\subsection{A5.Reliability Diagrams for ImageNet-C}

\begin{figure}[H]
    \centering
    \includegraphics[width=0.95\linewidth]{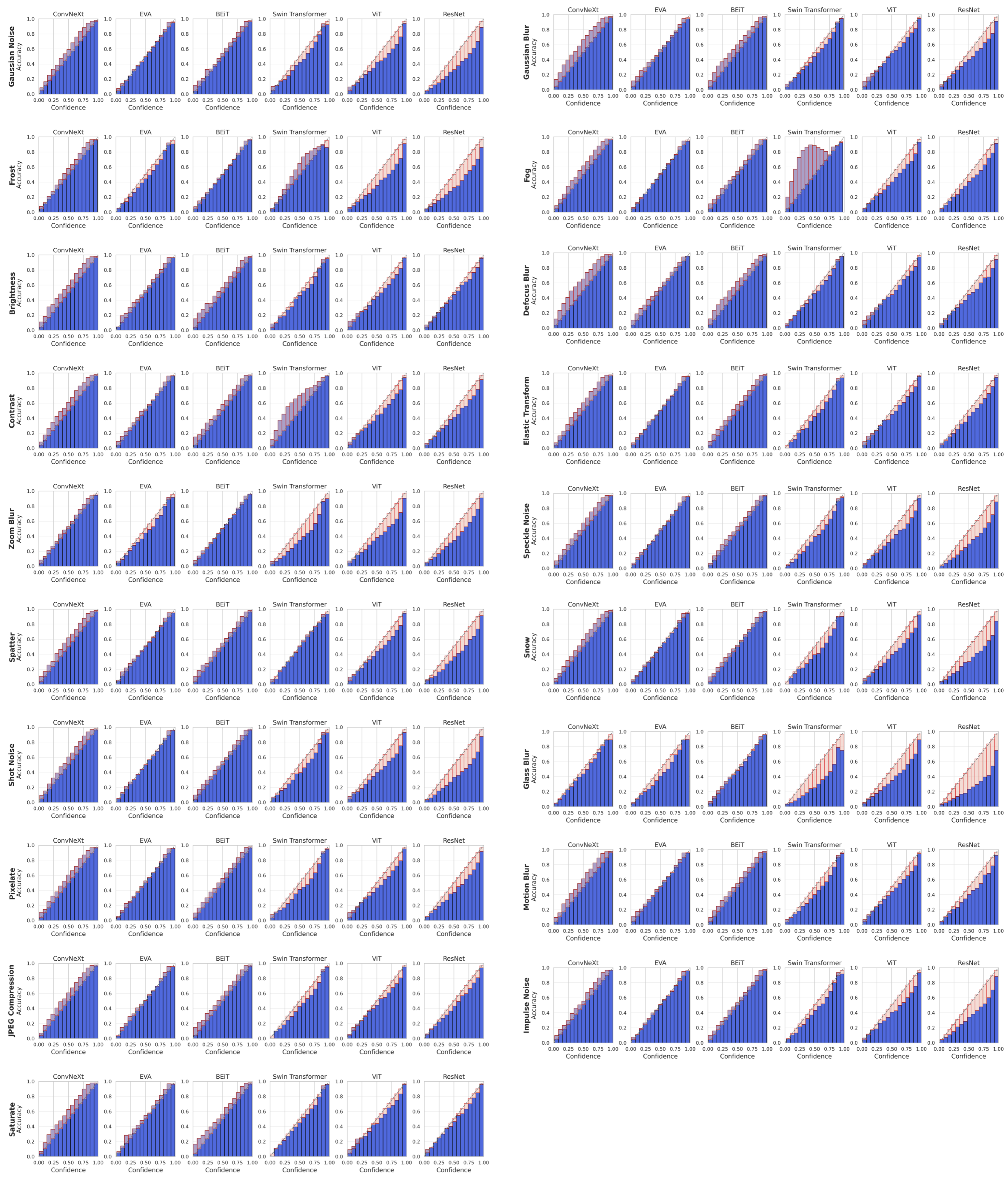}
    \caption{Individual reliability diagrams for all 19 synthetic corruptions of  ImageNet-C for severity 3.}
    \label{fig:reliability_corruption3}
\end{figure}

\begin{figure}[H]
    \centering
    \includegraphics[width=0.95\linewidth]{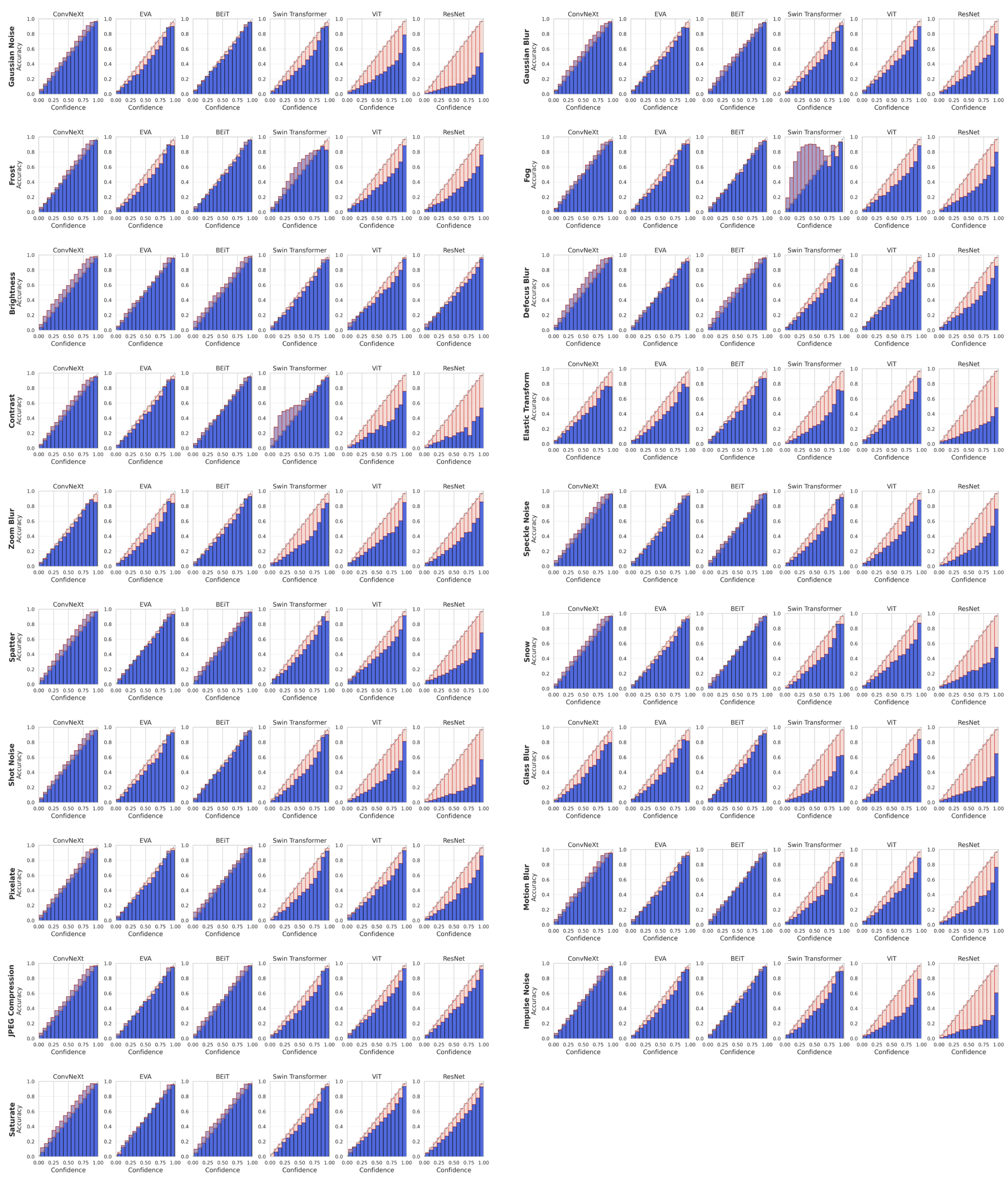}
    \caption{Individual reliability diagrams for all 19 synthetic corruptions of  ImageNet-C for severity 5.}
    \label{fig:reliability_corruption_5}
\end{figure}

\section{B: Visualization of Synthetic Perturbations in Dermatoscopic Imaging}

\begin{figure}[H]
    \centering
    \includegraphics[width=0.5\linewidth]{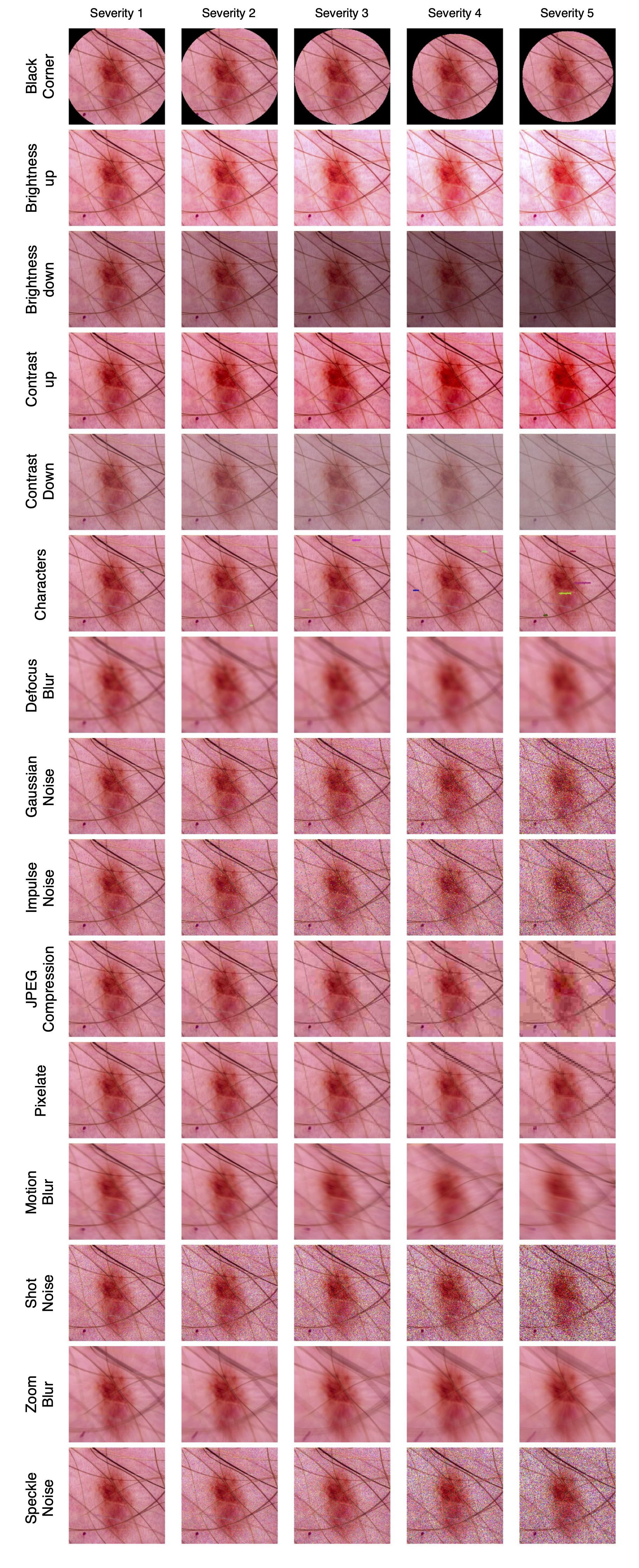}
    \caption{Example images of synthetic distribution shift for skin cancer classification task}
    \label{fig:derma_corruptions_example}
\end{figure}






\end{document}